\documentclass[lettersize,journal]{IEEEtran}
\usepackage{amsmath,amsfonts}
\usepackage{algorithmic}
\usepackage{algorithm}
\usepackage{array}
\pdfoutput=1
\usepackage{textcomp}
\usepackage{stfloats}
\usepackage{url}
\usepackage{verbatim}
\usepackage{graphicx}
\usepackage{cite}
\usepackage{subfigure}
\usepackage{color}
\usepackage{xcolor}
\usepackage{ragged2e}
\usepackage{threeparttable}
\usepackage{booktabs}
\usepackage{multirow}

\begin{document}

\title{Weakly-supervised Semantic Segmentation via Dual-stream Contrastive Learning of Cross-image Contextual Information}

\author{Qi Lai,
        Chi-Man Vong,~\IEEEmembership{Senior Member,~IEEE,}
        %and C.L. Philip Chen,~\IEEEmembership{Fellow,~IEEE}
        % <-this % stops a space
%\thanks{Q. Lai and C.M. Vong (\textit{Co-corresponding author}) are with the Department of Computer and Information Science, University of Macau, Macau 999078, China (Email: yb97447@um.edu.mo; cmvong@um.edu.mo).}% <-this % stops a space
%\thanks{C. L. Philip Chen is with the School of Computer Science and Engineering,South China University of Technology, Guangzhou 510006, China (e-mail:philip.chen@ieee.org)}% <-this % stops a space

\thanks{Manuscript received December 7, 2022.}}

%\markboth{IEEE Transactions on Neural Networks and Learning Systems,~Vol.~14, No.~8, December~2022}%
%{Shell \MakeLowercase{\textit{et al.}}: WSOD via interactive contrastive leanring of context information}

%\IEEEpubid{0000--0000/00\$00.00~\copyright~2022 IEEE}

\maketitle

\begin{abstract}
Weakly supervised semantic segmentation (WSSS) aims at learning a semantic segmentation model with only image-level tags. Despite intensive research on deep learning approaches over a decade, there is still a significant performance gap between WSSS and full semantic segmentation. Most current WSSS methods always focus on a limited single image (pixel-wise) information while ignoring the valuable inter-image (semantic-wise) information. From this perspective, a novel end-to-end WSSS framework called DSCNet is developed along with two innovations: i) pixel-wise group contrast and semantic-wise graph contrast are proposed and introduced into the WSSS framework; ii) a novel dual-stream contrastive learning (DSCL) mechanism is designed to jointly handle pixel-wise and semantic-wise context information for better WSSS performance. Specifically, the pixel-wise group contrast learning (PGCL) and semantic-wise graph contrast learning (SGCL) tasks form a more comprehensive solution. Extensive experiments on PASCAL VOC and MS COCO benchmarks verify the superiority of DSCNet over SOTA approaches and baseline models.
\end{abstract}

\begin{IEEEkeywords}
weakly supervised semantic segmentation; contrastive learning; dual-stream framework; cross-image contextual information.
\end{IEEEkeywords}

%------------------------------------------------------------------------
\section{Introduction}
\label{sec:intro}
\begin{figure}[!ht]
    \centering
        \subfigure[Only pixel-wise contextual information is utilized to update the pseudo segmentation labels.]{\includegraphics[width=2.8in]{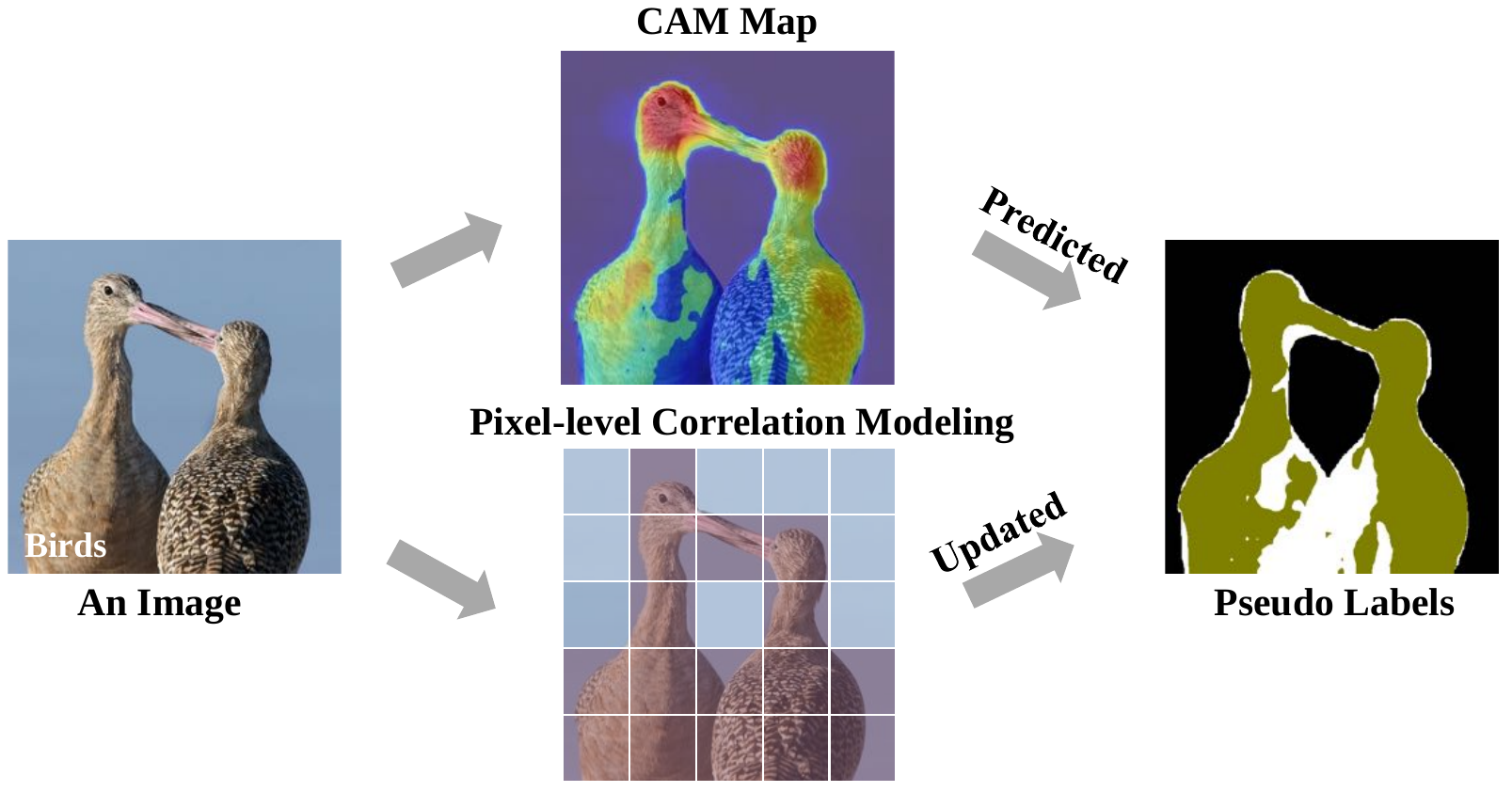}}
        %\label{fig:Fig_1a} \\
        \subfigure[Only semantic-wise contextual information is considered.]{\includegraphics[width=3in]{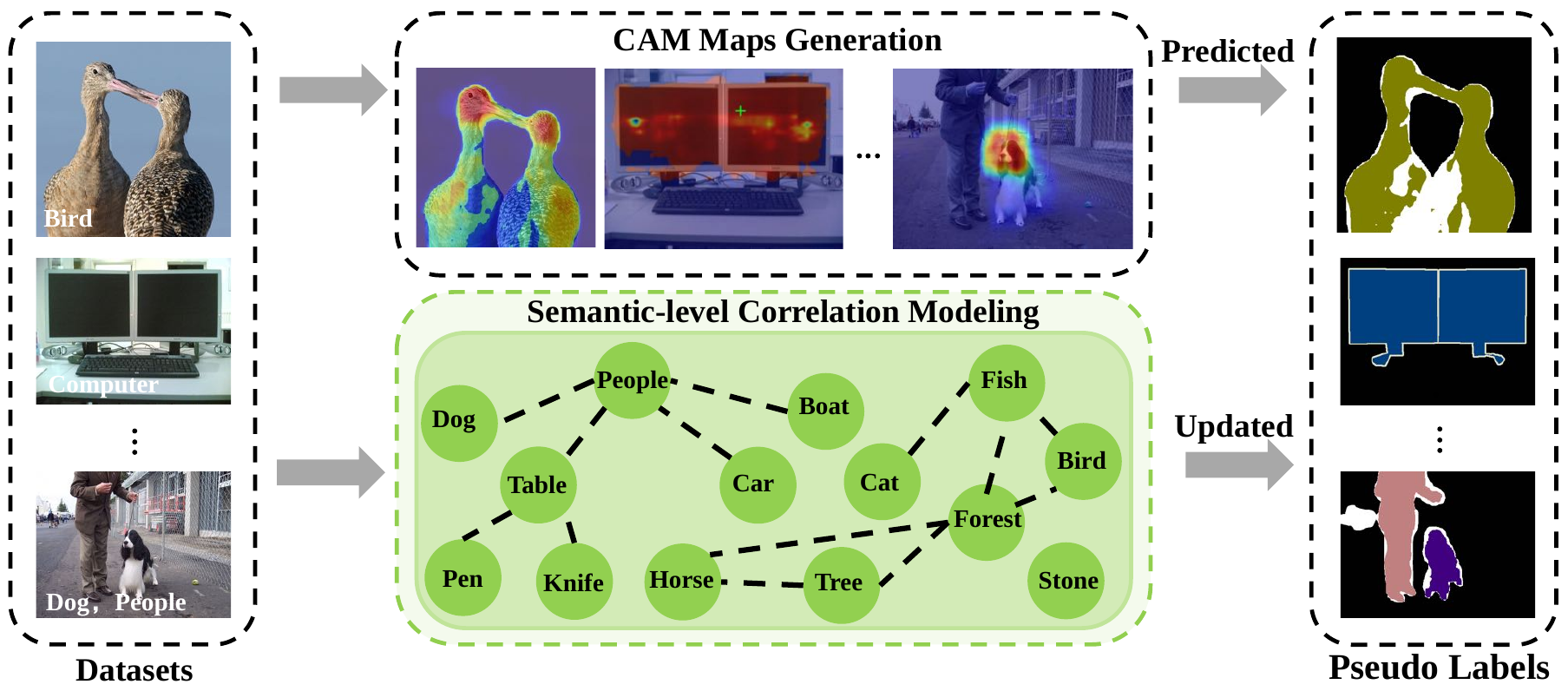}}
        %\label{fig: Fig_1b}\\
        \subfigure[Jointly learning dual contextual information, i.e., semantic- and pixel-wise contextual information]{\includegraphics[width=3in]{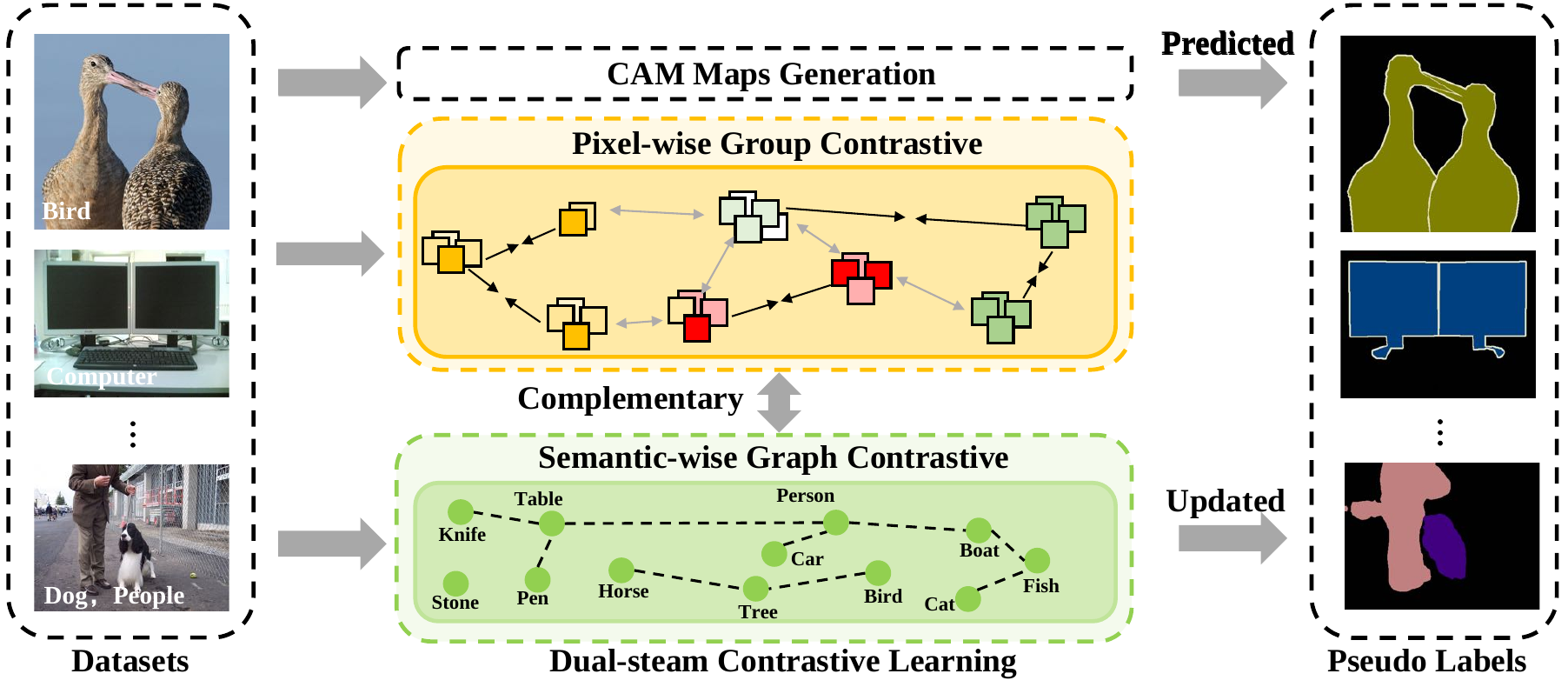}}
        %\label{fig: Fig_1c}\\
\label{fig:fig1}
\caption{Illustration of the difference between the existing works and our proposed method for weakly supervised semantic segmentation (WSSS) task. Our approach \textit{jointly} learns two types of cross-image contextual information (i.e., pixel-wise and semantic-wise contextual information) using only image-level ground-truth labels, and assists to generate the updated pseudo segmentation labels, providing more accurate supervision for semantic segmentation.}
\end{figure}

\begin{figure*}[ht]
\centering
\includegraphics[width=6.2in]{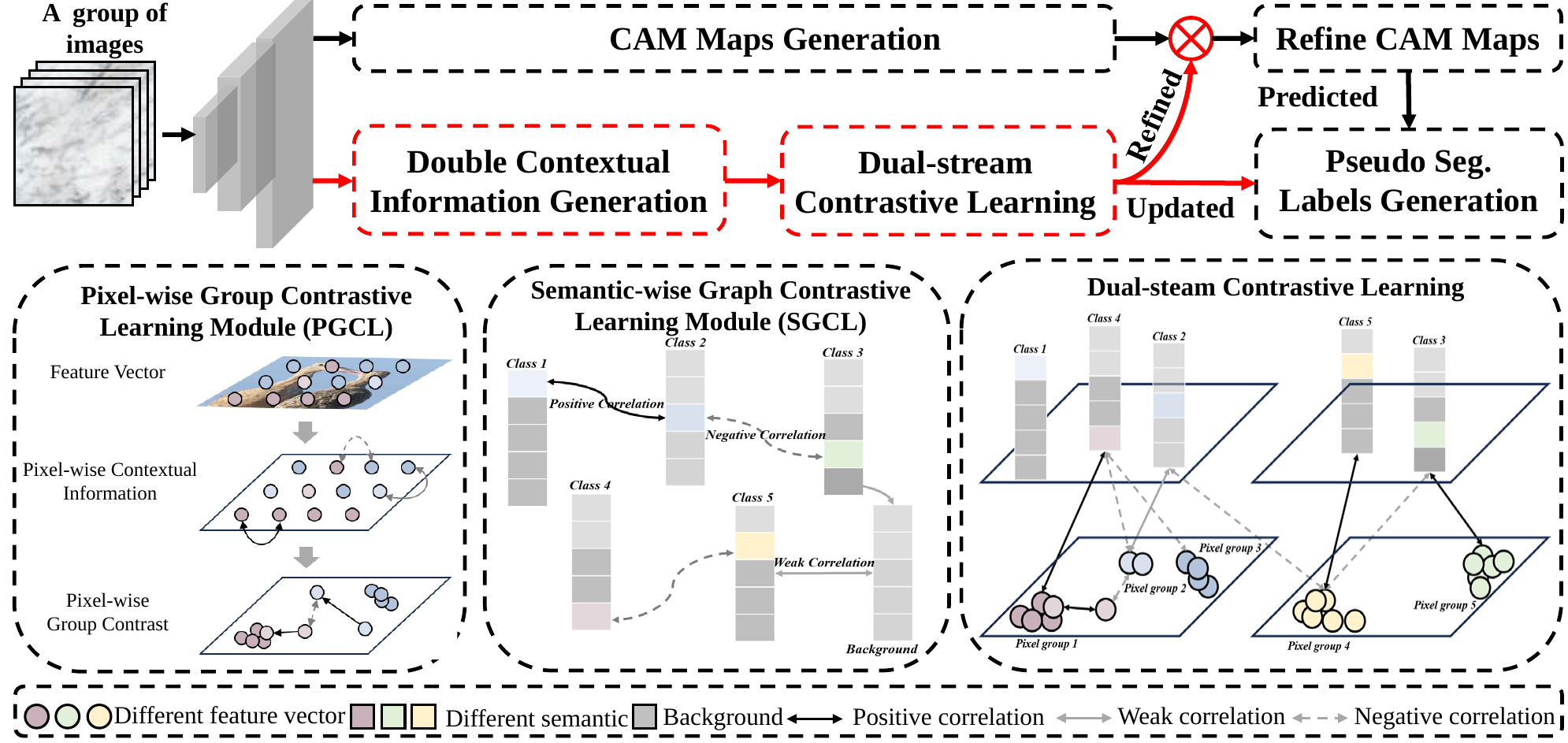}
\caption{An overview of the proposed DSCNet. Unlike the existing WSSS methods which use an RGB image as input, the input of DSCNet is a group of images.The “CAM” (class activation mapping) was first proposed in \cite{zhou2016learning}, which is a technique commonly used in weakly supervised semantic segmentation (WSSS) tasks \cite{lee2021bbam,zhou2021group,li2021pseudo,jiang2022l2g,sun2020mining,xu2021leveraging}. The “SGCL” and “PGCL” represent the semantic-wise graph contrast learning module and the pixel-wise group contrast learning module respectively. Note that the red lines represent the newly proposed module in this paper.}
\label{fig:workflow}
\end{figure*}

\IEEEPARstart{S}{emantic} segmentation \cite{li2021pseudo,jiang2022l2g,sun2020mining} describes the process of predicting and assigning semantic labels for each pixel of an input image. Obviously, training a fully supervised semantic segmentation (FSSS) model needs a lot of manpower and financial resources to annotate the huge amount of training images pixel by pixel. To alleviate this issue, low-cost annotation methods, e.g., scribbles \cite{lin2016scribblesup}, bounding boxes \cite{lee2021bbam}, points, and image-level labels \cite{zhou2021group,li2021pseudo}, have become an inevitable trend for semantic segmentation task, which is also known as weakly-supervised semantic segmentation (WSSS) \cite{lee2021bbam}. Among them, image-level labels are the easiest to obtain but also the most difficult to process. This is because the image-level labels only provide the category information of an image, while the location information of the object regions in the image is missing.

Most current WSSS approaches \cite{lee2021bbam,zhou2021group,li2021pseudo,jiang2022l2g,sun2020mining,xu2021leveraging} follow a two-step framework, i.e., generating pseudo segmentation labels (called \textit{pseudo labels} hereafter) and then training segmentation models. In evidence, the key to this framework is to generate high-quality pseudo labels that can better support segmentation model training. The groundbreaking work generates pseudo labels by developing \textit{class activation mapping} (CAM) \cite{zhou2016learning} which can mine the object regions from an activated image classifier. Although CAM can recognize object regions according to specific categories, these regions are very sparse with rough boundaries because the classifier can only activate a small part of features with strong discriminative capability. In order to obtain higher-quality pseudo labels, researchers have made many attempts, such as leveraging auxiliary significance supervision \cite{xu2021leveraging}, discriminative attention mechanism \cite{wu2021embedded,ru2022learning}, category semantic exploration \cite{sun2020mining,chang2020weakly}, or transform learning \cite{jiang2022l2g,xu2022multi}. However, all of these methods only focus on employing single image information (i.e., \textit{pixel-wise} contextual information) to acquire pseudo labels (Figure \ref{fig:fig1}(a)) while ignoring the valuable \textit{semantic-wise} contextual information among the image set. Recent studies proposed a group-wise learning framework \cite{zhou2021group} to capture the cross-image contextual information (Figure \ref{fig:fig1}(b)) but it only considers semantic-wise information while still ignoring pixel-level contextual information, resulting in limited accuracy to a certain extent. 

In summary, \textit{none} of the existing WSSS works can \textit{simultaneousl}y handle the two types of cross-image contextual information (i.e., both pixel-wise and semantic-wise), which can more accurately predict the pseudo labels. Moreover, most current WSSS approaches heavily depend on pseudo labels in the absence of proper supervision \cite{zhou2022regional}. This is not conducive to sufficiently understanding the entire image set and correctly building the overall correspondence from the semantic-level (or image-level label) to the pixel-level.

Motivated by the above observation, we introduce two types of contextual information into the WSSS task to maximize the use of context knowledge in the image set, as shown in Figure \ref{fig:fig1}(c). The purpose is to accurately establish the correspondence from semantic-level (image-level) to pixel-level and effective pseudo-label reasoning. 

Since the contextual information is under a weakly supervising scenario where only image-level labels are available, self-supervised learning method is necessary. Following the unprecedented success of \textit{contrastive learning} on self-supervised learning, contrastive learning is adopted to learn the two types of contextual information in our WSSS task. However, one technical difficulty arises, existing contrastive learning methods either ignore all annotations \cite{wang2021dense} or require pixel-level supervision \cite{zhong2021pixel}. Moreover, they can only \textit{independently} handle one type of contextual information (i.e., either pixel-wise or each semantic-wise contextual information, not both) but practically these two types of contextual information must be jointly learned. This challenging issue remains unresolved because it is not trivial to jointly optimize the dual contextual information.

To tackle this challenging issue, a novel dual-stream end-to-end WSSS framework (Figure \ref{fig:workflow}) is designed and called \textit{\textit{dual-stream contextual information contrastive learning network}} (DSCNet). In contrast to current WSSS works which independently consider either pixel-wise or semantic-wise contextual information, our proposed DSCNet exploits double contextual information from two novel perspectives:

\noindent \textbf{Pixel-wise group contrastive} tends to be group contrastive learning, that is, self-mining a group of positive pixel pairs for each category, rather than contrasting pixel-by-pixel. Compared with single-pixel contrastive, pixel groups not only ensure efficiency but are also able to express more complete semantic information.

\noindent \textbf{Semantic-wise graph contrastive} enables the model to learn to distinguish all possible object regions in the dataset, thus facilitating the establishment of a more comprehensive semantic-to-pixel correspondence. For example, if we know that a certain object always appears in a certain context, such as a tree appearing in an outdoor scene, then we can use this contextual information to help the model learn to segment the object more accurately.

\begin{table*}[!ht]
\centering
\caption{ The specific details of the comparison weakly supervised semantic segmentation (WSSS) methods.} \label{tab:wsss methods}
\resizebox{\linewidth}{!}{
\begin{tabular}{lccl}
\hline
\hline\noalign{\smallskip}
\multirow{2}{*}{Method}  & \multicolumn{2}{c}{Contextual information} &\multirow{2}{*}{Key Techniques} \\ \cline{2-3} 
                                             & Pixel-wise     & Semantic-wise          \\ \hline
OAA+ {[}ICCV19{]} \cite{jiang2019integral}   & $\surd$        &  &
\begin{tabular}[c]{@{}l@{}}• Design an Online Attention Accumulation (OAA) strategy;\\    
• Maintain a cumulative attention map for each target category in each training image.
\end{tabular}\\  \hline
CIAN {[}AAAI20{]} \cite{fan2020cian}         &$\surd$         &   &
\begin{tabular}[c]{@{}l@{}}• Proposed an end-to-end cross-image affinity module;\\
• Exploits pixel-level cross-image relationships.
\end{tabular}\\ \hline
Luo et.al{[}AAAI20{]} \cite{luo2020learning} &          & $\surd$&
\begin{tabular}[c]{@{}l@{}}• Explore the generic features across images from the same semantic category to discover the underlying structures;\\
• Design a self-contained and saliency-free segmentation system.
\end{tabular}\\ \hline
BES {[}ECCV20{]} \cite{chen2020weakly}       &   -      &  -  &
\begin{tabular}[c]{@{}l@{}}• Explore object boundaries;\\
• Using attention pooling to improve the performance of CAM.
\end{tabular}\\ \hline
\textit{Chang et.al}{[}CVPR20{]} \cite{chang2020weakly} & &$\surd$ &
\begin{tabular}[c]{@{}l@{}}• Via self-supervised learning to enhance feature representation to improve the initial CAMs for WSSS.
\end{tabular}\\ \hline
CONTA {[}NeurIPS20{]} \cite{zhang2020causal} &          & $\surd$  &
\begin{tabular}[c]{@{}l@{}}• A structural causal model to analyze the causalities among images, contexts, and class labels.
\end{tabular}\\ \hline
ICD {[}CVPR20{]} \cite{fan2020learning}      &          &$\surd$ &
\begin{tabular}[c]{@{}l@{}}• Proposed an end-to-end Intra-Class Discriminator (ICD);\\
• Learning an intra-class boundary to separate foreground objects and the background.
\end{tabular}\\ \hline
CPN {[}ICCV21{]} \cite{zhang2021complementary}& $\surd$ &  &
\begin{tabular}[c]{@{}l@{}}• Propose a Complementary Patch (CP) representation to enlarge the seed regions in CAM;\\
• Present a triplet network (CPN) with Triplet CP (TCP) loss and CP Cross Regularization (CPCR); \\
• A Pixel Region Correlation Module (PRCM) is proposed to further refine the CAM.
\end{tabular}\\ \hline
PMM {[}ICCV21{]} \cite{li2021pseudo}         & -        & -   &
\begin{tabular}[c]{@{}l@{}}• Proposed Proportional Pseudo-mask Generation with Coefficient of Variation Smoothing; \\
• Pretended Under-fitting Strategy, to generate high-quality pseudo-masks.
\end{tabular}\\ \hline
GroupWSSS {[}TIP21{]} \cite{zhou2021group}   & $\surd$  &$\surd$  &
\begin{tabular}[c]{@{}l@{}}• Propose group-wise semantic mining; \\
• Design a graph-aware solution to discover comprehensive semantic context from a group of images within an\\ effective iterative reasoning process.
\end{tabular}\\ \hline
EDAM {[}CVPR21{]} \cite{wu2021embedded}      &          &$\surd$ &
\begin{tabular}[c]{@{}l@{}}• Designed a Discriminative Activation (DA) layer; \\
• Using collaborative multi-attention module to explore the intra-image and inter-image homogeneity.
\end{tabular}\\ \hline 
SEAM {[}CVPR20{]} \cite{wang2020self}        & $\surd$  &    &
\begin{tabular}[c]{@{}l@{}}• Propose a self-supervised equivariant attention mechanism (SEAM); \\
• Incorporating equivariant regularization with pixel correlation module (PCM).
\end{tabular}\\ \hline 
AuxSegNet {[}ICCV21{]} \cite{xu2021leveraging} & $\surd$&$\surd$  &
\begin{tabular}[c]{@{}l@{}}• Leverages multi-label image classification and saliency detection as auxiliary tasks to help learn the WSSS task; \\
• Considering pixel-wise and semantic-wise contextual information independently.
\end{tabular}\\ \hline 
{ASDT {[}TIP23{]} \cite{zhang2023weakly}} & -       &  -  &
\begin{tabular}[c]{@{}l@{}}• Introducing two critical information into WSSS tasks, i.e., the discriminative object part and full object region by \\ using knowledge distillation; \\
 Building an end-to-end learning framework, alternate self-dual teaching, based on a dual-teacher single-student \\ network architecture.
\end{tabular}\\ \hline
DSCNet          & $\surd$       & $\surd$   &
\begin{tabular}[c]{@{}l@{}}• Exploiting two cross-image contextual information (pixel- and semantic-wise) among training data; \\
• Design a dual-stream contrastive learning framework to update pseudo labels by jointly learning pixel-wise and \\semantic-wise contextual information.
\end{tabular}\\ \hline 
\hline\noalign{\smallskip}
\end{tabular}
}
\end{table*}

These two context-information contrastive learnings are indispensable to our model. The semantic-wise graph contrastive helps the network to understand more structured input from a holistic view, while the pixel-wise group contrastive focuses on improving the positioning ability of the model over the object regions by comparing different pixel groups. Specifically, our proposed method includes two novel modules, i.e., \textit{Semantic-wise Graph Contrastive Learning module} (SGCL) and \textit{Pixel-wise Group Contrastive Learning module} (PGCL). SGCL exploits the semantic similarity to tap latent correspondence between image-level labels and each pixel, in view of the graph learning ability of GCN. Additionally, DSCNet is flexible and can be easily integrated into the existing WSSS model. Compared to the state-of-the-art WSSS models \cite{zhou2021group,xu2021leveraging}, DSCNet achieves consistently improved performance on PASCAL VOC 2012 \cite{everingham2009pascal} and COCO 2014 \cite{lin2014microsoft}. 

In summary, the \textbf{contributions} of this paper are as below:

\noindent i) DSCNet solves an essential yet long ignored problem in WSSS by \textit{simultaneously} exploiting two helpful types of contextual information (pixel-wise and semantic-wise) among training data. This significantly helps construct the correspondence between image-level semantic concepts and pixel-level object regions.

\noindent ii) Technically, a novel dual-stream contrastive learning framework is newly designed which is able to update pseudo labels by learning pixel-wise and semantic-wise contextual information, so as to continuously improve the semantic segmentation performance. 
%----------------------------------------------------------------------------------
\section{Preliminaries}
\label{sec:Preliminaries}
\subsection{Weakly-Supervised Semantic Segmentation (WSSS)} 
\label{ssec:wsss}
\textit{Weakly-Supervised Semantic Segmentation} (WSSS)\cite{xu2021leveraging,wu2021embedded,ru2022learning} can be defined as the problem of learning a semantic segmentation model under limited or weak supervision. WSSS is increasingly popular for its practical value in reducing the burden of the large-scale collection of pixel-level annotations required by fully supervised situations, especially in special scenarios, e.g., medical image segmentation with imperfect annotations \cite{fang2023reliable}.The key challenge of WSSS is to train a model with only image-level labels that produce more accurate pseudo-segmentation labels to guide the subsequent semantic segmentation task.
%Formally, let $\mathcal{X}$ be the input image set and $\mathcal{Y}$ be its corresponding image-level label. 
For this reason, most efforts are now devoted to obtaining high-quality pseudo labels, such as discriminative attention mechanism \cite{wu2021embedded,ru2022learning}, category semantic exploration \cite{sun2020mining,chang2020weakly}\cite{zhang2022generalized}, or transfer learning-based methods \cite{jiang2022l2g,xu2022multi}, as listed in Table \ref{tab:wsss methods}. As we can see, most existing WSSS methods focus on considering only one type of contextual information, i.e., either pixel-wise \cite{jiang2019integral,fan2020cian,zhang2021complementary,wang2020self}, or semantic-wise \cite{luo2020learning,chang2020weakly,zhang2020causal,fan2020learning}. Among them, only GroupWSSS  \cite{zhou2021group} and AuxSegNet \cite{xu2021leveraging} have considered the contextual information between pixel- and semantic-wise, but they use pixel by pixel scheme to capture pixel-level correlation (see Figure \ref{fig:PGCL} (a)), which is very time-consuming. In addition, both of them \textit{independently} consider these two contextual information, which is fundamentally different from our DSCNet of refining CAM segmentation results through both contextual information \textit{simultaneously}. For this reason, DSCNet is designed to simultaneously consider these two types of contexts (both pixel-wise and semantic-wise) by constructing pixel groups, which can i) learn the dependencies between pixels; ii) ensure high efficiency; and iii) express more complete semantic information. Furthermore, a new loss function for joint optimization is designed. Both these two contextual information complement each other to achieve better segmentation results.

\subsection{Contrastive Learning} 
\label{ssec:contrastive}
\textit{Contrastive learning} is a type of self-supervised machine learning technique that aims to learn a function that can distinguish between pairs of similar and dissimilar data points. Except used in image or instance-level discrimination on classification tasks, current efforts \cite{wang2021dense,zhong2021pixel,alonso2021semi} exploit pixel-level or patch-level discrimination as pretext task in order to pre-train a model and then guide or fine-tune for the denser downstream tasks (e.g., semantic segmentation or instance segmentation). Notably, contrastive learning \cite{zhong2021pixel,alonso2021semi} was introduced into semi-supervised semantic segmentation which extends the self-supervised setting (by using label information) and compares the set of the labeled samples as positives and against the unlabeled samples as negatives. Inspired by \cite{zhong2021pixel,alonso2021semi}, our method adopts weakly supervised annotations to perform dense contrastive learning to enhance the ability of CAM to locate the “target regions”. Yet, existing dense contrastive learning approaches either ignore all annotations \cite{koshkina2021contrastive,wang2021dense} or require pixel-level supervisions \cite{zhong2021pixel,chaitanya2020contrastive}, which does not meet the requirements of the WSSS task. In addition, contrastive learning at the pixel-level is very time-consuming, which seriously affects the efficiency of the model \cite{zhou2021group}.

%============================================
\section{Methodology}
\label{sec:methodology}
In this section, we will introduce DSCNet in detail, including problem definition, pixel-wise group contrastive learning (PGCL) module, semantic-wise graph contrastive learning (SGCL) module, dual-stream contrastive learning (DSCL) mechanism, and some details of the network architecture.

\subsection{Problem definition}
\label{sbusec:pro_def}

The overview of our proposed DSCNet is shown in Figure \ref{fig:workflow}. Unlike current approaches that focus on single image information or \textit{independently} consider two types (pixel-wise, and semantic-wise) of contextual information (see Figure \ref{fig:fig1}(a) and \ref{fig:fig1}(b)), our proposed DSCNet can \textit{simultaneously} mine both pixel-wise and semantic-wise contextual information from a group of images through the newly designed PGCL and SGCL.
%through our proposed dual-stream contrastive learning mechanism
In this way, DSCNet can mitigate the incomplete annotation issue in WSSS and generate high-quality pseudo labels. Formally, each training image $X_i \in \mathbb{R}^{w\times h \times 3}$ in the image set $\mathcal{X}$ corresponds to image-level label $Y_i \in \left \{ 0,1 \right \} ^K$ only, where $K$ is the number of categories. In this paper, we follow the general way [5-8, 10-13] to generate the CAM maps and the pseudo-segmentation labels. That is, the image set $\mathcal{X}=\left \{ X_i \right \}_{i=1}^N $ first passes through a backbone network, e.g., ResNet38 [34] to extract feature map $F=\left \{ f_i \right \}_{i=1}^N$, where $f_i \in \mathbb{R}^{(WH)\times D}$ is the dense feature embedding of the $i$-th image, with $W\times H$ spatial size and $D$ channels. $N$ denotes the total number of the training image. Then the class-aware attention map of the \textit{i}-th image $O_i$ is generated as below:
\begin{equation}
  O_i = \mathcal{F}_{CAM}(f_i)
  \label{eq:cam}
\end{equation}
where $O_i= [ O_{i,1},O_{i,2},…,O_{i,K}]  $, and each $ O_{i,k}\in \mathbb{R}^{W \times H}$ represents the CAM map for $k$-th category. $\mathcal{F}_{CAM}$ represents class-aware convolutional layer. 

After generating the CAMs $O$ of each image, a pixel-wise argmax function was employed to obtain the pseudo segmentation labels $y_i$,
\begin{equation}
  y_i = argmax(O_i)
  \label{eq:y}
\end{equation}
where $y_i \in \mathbb{R}^{W \times H}$ associates each pixel with category. Note that only image-level ground-truth labels are required to train the proposed DSCNet.

%-------------------------------------------------------------------------
\subsection{Pixel-wise Group Contrastive Learning Module (PGCL)}
\label{sbusec:PGCL}
Existing WSSS methods \cite{zhang2021complementary,zhou2022regional,xu2021leveraging} have shown that pixel-wise contextual information modeling is very important. However, current WSSS approaches still suffer from several drawbacks: i) time-consuming training procedure; ii) only focus on the pixel-wise contextual information in a single image without considering the semantic-wise contextual information between pixels across images. These two problems were rarely noticed in the existing literature and only a few methods \cite{wang2021exploring} tried to model the pixel-to-semantic contextual information. A classic and common method is to model the contextual information pixel-by-pixel by using contrastive learning \cite{grill2020bootstrap}, 
\begin{equation}
  \mathcal{F}_{Con}(f_{i,u},f_{i,u}^+)=-log \frac{\varphi (f_{i,u},\frac{f_{i,u}^+}{\tau })}{ {\sum_{f_{i,u}^- \in \mathcal{N}_i}}\varphi(f_{i,u},\frac{f_{i,u}^-}{\tau }) }
  \label{eq:con_p}
\end{equation}
where $\mathcal{F}_{Con}$ indicates the contrastive learning function \cite{grill2020bootstrap}. $f_{i,u}$  denotes the $u$-th pixel in the $i$-th image and the $f_{i,u}^+$ represents this pixel belongs to positive. $\mathcal{N}_i$ contains all pixels that are negative. $\tau>0$ represents a temperature hyperparameter. 
And $\varphi(,)$  denotes the \textit{cosine similarity} dot product, i.e., $\varphi(u,v) = exp(sim(u,v)=exp(\frac{u \cdot v}{\left \| u \right \|_2 \left \| v \right \|_2} )$,  where $\left \| \cdot \right \|_2$ indicates $l_2$-normalized.
%$\varphi(u,v) = exp(sim(u,v))$, $sim(u,v)$ denoets the \textit{cosine similarity} dot product, i.e., $sim(u,v)=\frac{u \cdot v}{\left \| u \right \|_2 \left \| v \right \|_2} $,} where $\left \| \cdot \right \|_2$ indicates $l_2$-normalized.

\begin{figure}[!ht]
    \centering
        \subfigure[Traditional pixel-wise contextual information contrastive. The model performs contrast pixel-by-pixel.]{\includegraphics[width=3.3in]{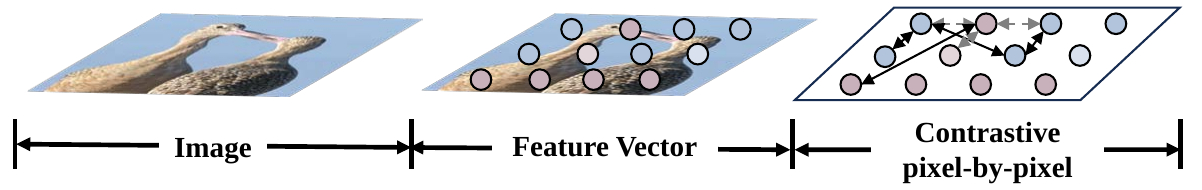}}
        %\label{fig:Fig_1a} \\
        \subfigure[Our proposed pixel-wise group contrastive learning (PGCL) module performs contrast in the group.]{\includegraphics[width=3.3in]{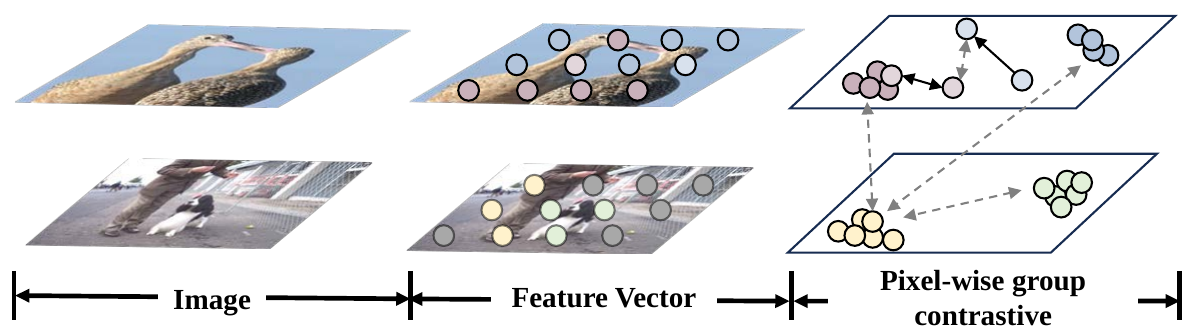}}
        %\label{fig: Fig_1b}\\
%\includegraphics[width=3.3in]{Figure/Fig_PGCL.pdf}
\caption{A brief illustration of the difference between traditional pixel-wise contrastive mechanism and our proposed pixel-wise group contrastive learning (PGCL) module. The colored circles represent the feature vector belonging to different categories. The double arrows of black, gray, and gray dashed lines respectively represent "positive correlation", "weak correlation", and "negative correlation".}
\label{fig:PGCL}
\end{figure}

Nevertheless, there are two issues with this pixel-by-pixel modeling method: 1) it is very time-consuming, especially for an image of high resolution. 2) In addition, it is difficult for a single pixel to accurately express semantic information. Thus, we design a pixel-wise group contrastive learning (PGCL) module to tackle these two issues. As illustrated in Figure \ref{fig:PGCL} (b), pixel-wise contextual information is used as the guiding information to complete the clustering of pixels, and then apply contrastive learning on each pixel group. 

The pixel-wise contextual information between each pixel can be represented as follows:
\begin{equation}
  \mathcal{C}_i,v = \sum_{v=1}^V cov (f_{i,u}, f_{i,v})
  \label{eq:pwci}
\end{equation}
where $\mathcal{C}_i,v \in \mathbb{R}^D$ represents the contextual information between the $v$-th pixel and other neighboring pixels in $i$-th image while $V$ is the total number of the pixel in the $i$-th image. $cov(\cdot)$ indicates the \textit{Pearson correlation coefficient} function \cite{cohen2009pearson}, which is determined by the similarity between two pixel, e.g., $f_{i,u}$ and $f_{i,v}$. $f_{i,u}$ and $f_{i,v}$ denote the $u$-th and $v$-th pixel in the $i$-th image, respectively. In Eq. (\ref{eq:pwci}), the purpose of constructing pixel-wise contextual information is to learn a feature embedding, by clustering the pixels with similar characteristics and by pushing dissimilar samples apart. 
\begin{equation}
  p_{i,u} =  \frac{1}{\left | Y_i \right | } \sum_{u=1}^{\left | Y_i \right | } \frac{1}{\left | \mathcal{P} \right | } \sum_{f_{i,u} \in \mathcal{P}} f_{i,u} 
  \label{eq:piu}
\end{equation}
where $p_{i,u}$ denotes the $u$-th \textit{disjoint group} in the $i$-th image, i.e., $ p_{i,u}\cap_{u\neq v}p_{i,v}=\emptyset $  with  ${\textstyle\bigcup_{u=1}^{\left | Y_i \right |}} p_{i,u}=f_i$ by \textit{k-means algorithm} \cite{yuan2019research}. Note that the number of the disjoint groups is determined by the number of categories $\left | Y_i \right |$ contained in each image and hence each image has $\left | Y_i \right |$ disjoint groups. $\mathcal{P} = \left \{ f_{i,u}\in X_i | \mathcal{C}_{i,u} < 0.5 \right \} $ indicates the sample set positively related to $f_{i,u}$. Similarly, $\mathcal{N}$ represents a negatively correlated sample set.

\begin{figure}[!ht]
\centering
\includegraphics[width=3.2in]{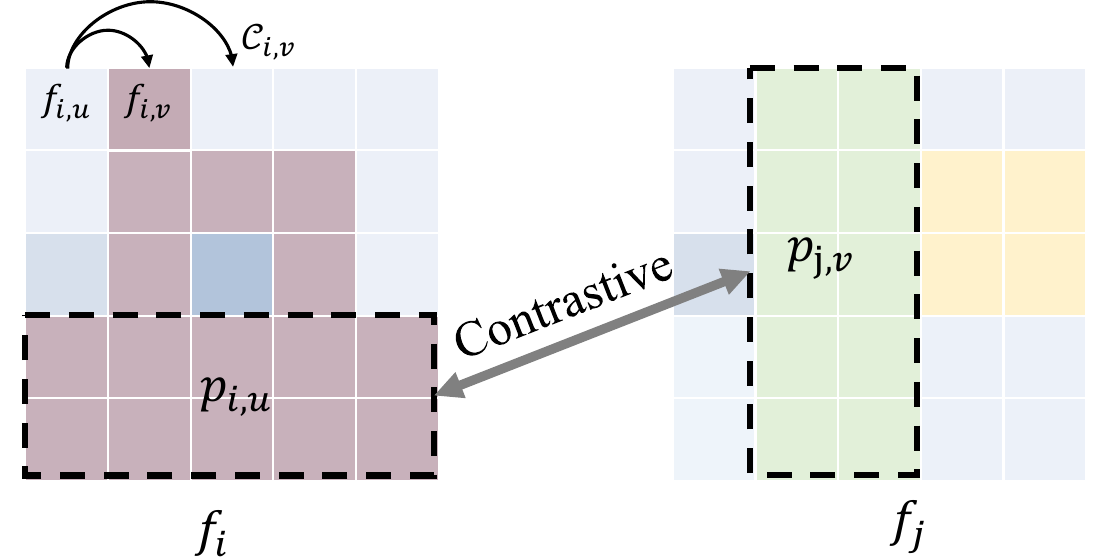}
\caption{Details of the pixel-wise group contrast.}
\label{fig:fig_4}
\end{figure}

For the PGCL module, by regarding each pixel embedding group as an individual object, the feature embedding $f_i$ of each image is divided into $\left | Y_i \right |$ disjoint groups $\left \{ p_{i,u} \right \}_{u=1}^{\left | Y_i \right |}$. Then, the pixel-wise group contrastive learning can be reformulated as below:
\begin{equation}
  \mathcal{F}_{Con}(p_{i,u},p_{j,v})=-log \frac{\varphi (p_{i,u},\frac{p_{j,v}}{\tau })}{ {\sum_{p_{j,v} \in \mathcal{N}_i}}\varphi(p_{i,u},\frac{p_{j,v}}{\tau }) }
  \label{eq:con_group}
\end{equation}
where $\mathcal{F}_{Con}(\cdot)$ represents the contrast function, $p_{i,u}$ and $p_(j,v)$ denote different pixel groups in the feature embedding space, see Figure \ref{fig:fig_4}. Note that $i=j$ and $u=v$ cannot establish at the same time.

With the definition of pixel-wise group contrastive learning in Eq. (\ref{eq:con_group}), the contrastive loss of PGCL can be formulated as:
\begin{equation}
  \mathcal{L}_{PGCL}=\frac{1}{N}\sum_{i=1}^N \frac{1}{\left | Y_i \right |} \sum{ \frac{1}{\left | Y_i \right |} }\mathcal{F}_{Con}(p_{i,u},p_{j,v})
  \label{eq:pgcl_loss}
\end{equation}
where $N$ and $\left | Y_i \right |$ denote the total number of the samples and pixel groups in the $i-$th image, respectively. 
%-------------------------------------------------------------------------

\subsection{Semantic-wise Graph Contrastive Learning Module (SGCL)}
\label{sbusec:SGCL}

Semantic-wise contextual information can provide additional learning clues for accurate modeling, i.e., by leveraging semantic-wise context information, the model can learn to associate visual features with specific semantic categories more effectively. With the above analysis of current approaches \cite{zhang2021complementary,zhou2021group,zhou2022regional,xu2021leveraging}, it is found that they focus on exploiting the limited intra-image contextual information, causing difficulty to understand semantic patterns more comprehensively. To resolve these issues, a novel module SGCL is developed to learn the semantic-wise contextual information for semantic-wise contrastive learning, which is calculated as follows:
\begin{equation}
  \mathcal{S}_{i,u}^k = \frac{1}{\left | Y_i \right |} \sum_{u=1}^{\left | Y_i \right |} \frac{1}{K}\sum_{k=1}^{K} M_{i,u}^k \cdot p_{i,u}
  \label{eq:swci}
\end{equation}
where $\mathcal{S}_{i,u}^k$ denotes the semantic consistency for the $k$-th category and pixel group $p_{i,u}$. $M_{i,u}^k$ indicates the similarity matrix which measures the probability of the pixel group $p_{i,u}$ belonging to the $k$-th category. Following the same procedure in the pixel-wise group contrastive loss, the semantic-wise graph contrastive loss can be computed as
\begin{equation}
  \mathcal{L}_{SGCL}=\frac{1}{N}\sum_{i=1}^N \mathcal{F}_{Con}(\frac{1}{\left | Y_i \right |}\sum_{u=1}^{\left | Y_i \right |}\mathcal{S}_{i,u}^k, y_i )
  \label{eq:sgcl_loss}
\end{equation}
where $\mathcal{F}_{Con}$ indicates the contrast function in Eq. (\ref{eq:con_p}) while $y_i$ denotes the pseudo label of the $i-$th image, which is obtained by the original CAM maps by Eq. (\ref{eq:y}).

%------------------------------------------------------------------------
\subsection{Dual-stream Contrastive Learning (DSCL) Mechanism}
\label{ssec:DSCL mechanism}
The dual-stream contrastive learning framework uses two separate streams to learn the representations of input samples. These two streams are trained \textit{simultaneously} to maximize the contrast between samples from different classes while minimizing the contrast between samples from the same class. In DSCNet, the pixel- and semantic-wise contrast in Eqs. (\ref{eq:pgcl_loss}) and (\ref{eq:sgcl_loss}) are complementary to each other, as shown in Figure \ref{fig:workflow}. In particular, the pixel-wise group contrastive allows the model to capture discriminative pixel groups, and it can better associate pixels with regions rather than a single pixel. The semantic-wise graph contrast can help regularize the feature embedding space and improve intra-class compactness and inter-class separability by explicitly exploiting the global semantic-wise contextual information between pixel groups. Hence, for each image $X_i$, the feature embedding $\widetilde{f}_i$ can be refined as:
\begin{equation}
  \widetilde{f}_i = \left ( softmax\left (f_i\otimes \mathcal{C}_i^T \right )\otimes \mathcal{S}_i\right )
  \label{eq:updated_fea}
\end{equation}
where $\widetilde{f}_i\in \mathbb{R}^{(WH)\times D}$ denotes an enriched feature representation of $f_i$, and then reshaped into W×H×D. $softmax(\cdot)$ normalizes each row of the input. $\mathcal{C}_i\in \mathbb{R}^{\left | Y_i  \right | \times D} $ and  $\mathcal{S}_i\in \mathbb{R}^{\left | Y_i  \right | \times K} $ indicate pixel- and semantic-wise contextual information, respectively. $\otimes$ represents matrix multiplication. 

Concatenated $\widetilde{f}_i$ and the original feature $f_i$, we have $\hat{f}_i=[f_i,\widetilde{f}_i]$. From this way, $\hat{f}_i$ encodes the pixel- and semantic-wise contextual information, hence obtaining the richer representation of the input data.
And then the CAM maps can be refined according to the  enriched feature representation  $\hat{f}_i$,
\begin{equation}
  \hat{O}_i = \mathcal{F}_{CAM}(\hat{f}_i) \in \mathbb{R}^{W \times H \times K}
  \label{eq:updated_o}
\end{equation}
where $\hat{O}_i$ denotes the refined CAM map of the $i$-th image.
Finally, the pseudo labels can be updated according to the refined CAM maps, 
\begin{equation}
  \hat{y}_i=argmax( \hat{O}_i )
  \label{eq:updated_y}
\end{equation}
where $argmax(\cdot )$ indicates the a pixel-wise argmax function \cite{du2022weakly}.

%------------------------------------------------------------------------
\subsection{Detail Network Architecture}
\label{ssec:network arch}
Our proposed DSCNet has five major components (see Figure \ref{fig:workflow}). 

\noindent \textbf{i}) FCN encoder, which maps all training sample $\mathcal{X}$ into dense embeddings $F=\mathcal{F}_{FCN}(\mathcal{X})$. In DSCNet, any FCN backbones can be employed. According to the classical WSSS approaches, ResNet38 is used here.

\noindent \textbf{ii}) CAM maps generation, i.e., Eq. (\ref{eq:cam}). In our work, $\mathcal{F}_{FCN}$ is adopted to Eqs. (\ref{eq:cam}) and (\ref{eq:updated_o}) independently.

\noindent \textbf{iii}) Pixel-wise group contrast and semantic-wise graph contrast, respectively, i.e., Eqs. (\ref{eq:pgcl_loss}) and (\ref{eq:sgcl_loss}).

\noindent \textbf{iv}) Refined CAM maps.
The specific procedure for updating CAM and pseudo labels:
(a) obtaining the updated feature embedding $\hat{f}_i$ which encodes the pixel- and semantic-wise contextual information for richer representation; (b) updating the CAM maps by Eq. (\ref{eq:updated_o}); (c) updating the pseudo labels by Eq.(\ref{eq:updated_y}).

\noindent \textbf{v}) The overall objective function is defined as:
\begin{equation}
  \mathcal{L}_{DSCL} = \alpha  \mathcal{L}_{PGCL}+\beta\mathcal{L}_{SGCL}+\mathcal{L}_{CE}
  \label{eq:total_loss}
\end{equation}
Here, $\mathcal{L}_{DSCL}$ represents the overall loss function, which is a combination of three loss terms: $\mathcal{L}_{PGCL}$, $\mathcal{L}_{SGCL}$, and $\mathcal{L}_{CE}$. Among them, $\mathcal{L}_{PGCL}$ denotes the pixel-wise group contrastive learning loss. The second term $\mathcal{L}_{SGCL}$ is the semantic-wise graph contrastive learning loss, while the third loss is the main cross-entropy loss imposing on the refined CAM maps $\hat{O}$. The hyperparameters $ \alpha $ and $\beta$ control the relative importance of the pixel-wise and semantic-wise consistency losses, respectively. By minimizing the overall loss function, the network can simultaneously learn pixel- and semantic-wise contextual information that is consistent with each other.

%-----------------------------------------------------------------------------
\section{Experiments}
\label{sec:Experimental}
\subsection{Experimental Setup}
\label{sbusec:setup}
\textbf{Dataset.} Following the \cite{luo2020learning,li2021pseudo,zhou2022regional,du2022weakly}, our proposed method is verified on PASCAL VOC 2012 \cite{everingham2009pascal} and MS COCO 2014 \cite{lin2014microsoft} datasets.
\textbf{ PASCAL VOC} consists of 21 semantic categories with a background. The dataset was divided into three subsets training (1,464 images), validation (1,449 images), and test (1,456 images). \textbf{MS COCO} is much more challenging than the Pascal VOC dataset which contains 81 categories with background categories. We follow [41-43] to train on the standard training set (90K images) and validation set (40K images).

\textbf{Evaluation Metrics.} We follow previous work \cite{xu2021leveraging,zhou2022regional,du2022weakly}, which adopted the mean Intersection-over-Union (mIoU) to verify the performance of DSCNet on the validation and test sets on PASCAL VOC and the validation set of MS COCO. The IoU measures the overlap between the predicted segmentation map and the ground truth segmentation map. It is calculated as the ratio of the intersection of the two maps to their union.

%\textbf{Implementation Details.} In this work, we employ two strong models (i.e, SEAM \cite{wang2020self} and AuxSegNet \cite{xu2021leveraging}) as our baseline in DSCNet for the experiments which all employ ResNet38 as backbones. SEAM \cite{wang2020self} focuses on improving CAM maps to improve the segmentation performance by using a self-supervised equivariant attention mechanism, while AuxSegNet \cite{xu2021leveraging}) adopts multiple auxiliary tasks to further enhance the segmentation result. Since our proposed method also involves self-supervised learning and the idea of joint learning, we establish DSCNet based on these two baselines. The weights of the backbones are pre-trained on the training set of each dataset. During training, the mini-batch size for SGD is respectively set to be 4, and 32 for PASCAL VOC, and MS COCO. An initial learning rate is set to 1e-3 for the first 40, 15, and 85K iterations for the PASCAL VOC and MS-COCO datasets, respectively. For the hyper-parameter, we discuss it in Section \ref{sbusec:para_ana}. 

%------------------------------------------------------------------------
\subsection{Comparison Result}

\begin{table}[!ht]
\centering
\caption{ Comparison result of WSSS methods on PASACL VOC 2012 \textit{val} and \textit{test} sets. The best result and the improvements over the baseline model are marked in bold and green, respectively. \label{tab:voc}}
\resizebox{\linewidth}{!}{
\begin{tabular}{lccc}
\hline
\hline\noalign{\smallskip}
\multirow{2}{*}{Method} & \multirow{2}{*}{Backbone} & \multicolumn{2}{c}{mIoU $(\%)$} \\ \cline{3-4} 
                        &              & val           & test          \\ \hline
OAA+ {[}ICCV19{]} \cite{jiang2019integral}       & ResNet101    & 65.2          & 66.4          \\
CIAN {[}AAAI20{]} \cite{fan2020cian}      & ResNet101    & 64.3          & 65.3          \\
Luo et.al{[}AAAI20{]} \cite{luo2020learning}   & ResNet101    & 64.5          & 64.6          \\
BES {[}ECCV20{]} \cite{chen2020weakly}       & ResNet50     & 65.7          & 66.6          \\
\textit{Chang et.al}{[}CVPR20{]} \cite{chang2020weakly} & ResNet101    & 66.1          & 65.9          \\
CONTA {[}NeurIPS20{]} \cite{zhang2020causal}  & ResNet38     & 66.1          & 66.7          \\
ICD {[}CVPR20{]} \cite{fan2020learning}       & ResNet101    & 67.8          & 68            \\
CPN {[}ICCV21{]} \cite{zhang2021complementary}       & ResNet38     & 67.8          & 68.5          \\
PMM {[}ICCV21{]} \cite{li2021pseudo}       & ResNet38     & 68.5          & 69            \\
GroupWSSS {[}TIP21{]} \cite{zhou2021group}  & VGG16        & 68.7          & 69            \\
EDAM {[}CVPR21{]} \cite{wu2021embedded}      & ResNet38     & 70.9          & 70.6          \\ \hline
\hline\noalign{\smallskip}
SEAM {[}CVPR20{]} \cite{wang2020self}      & ResNet38     & 64.5          & 65.7         \\
DSCNet+SEAM               & ResNet38     & 67.7 \textcolor{green!80}{↑ 3.2}         & 68.5 \textcolor{green!80}{↑ 2.8}          \\ \hline
\hline\noalign{\smallskip}
AuxSegNet {[}ICCV21{]} \cite{xu2021leveraging}  & ResNet38     & 69            & 68.6          \\
DSCNet+AuxSegNet          & ResNet38     & 70.3 \textcolor{green!80}{↑ 1.3}          & 71.1 \textcolor{green!80}{↑ 2.5}         \\ \hline
\hline\noalign{\smallskip}
\end{tabular}
}
\end{table}

\textbf{PASCAL VOC.} Table \ref{tab:voc} reports the comparison result of DSCNet against state-of-the-art WSSS approaches on Pascal VOC \textit{val} and \textit{test} sets. Our proposed DSCNet improves the segmentation performance of two baselines (i.e., SEAM and AuxSegNet). As seen, DSCNet has 3.1\% and 3.8\% improvements on SEAM and brings consistent improvements (1.3\% and 2.5\%) for AuxSegNet. The visualization results on VOC\textit{ val} is shown in Figure 5. As we can see, our proposed method shows remarkable capabilities in various and complex scenes.

\begin{table}[!ht]
\centering
\caption{ Comparison result of WSSS methods on COCO 2014 \textit{val}.\label{tab:coco}}
\begin{tabular}{lcc}
\hline
\hline\noalign{\smallskip}
method                 & backbone  & mIoU $(\%)$ \\ \hline
IAL {[}IJCV20{]} \cite{wang2020weakly}      & VGG16     & 20.4      \\
GroupWSSS {[}TIP21{]} \cite{zhou2021group} & VGG16     & 28.7      \\
\textit{Luo et.al} {[}AAAI20{]} \cite{luo2020learning} & ResNet101 & 29.9      \\
ADL {[}TPAMI20{]} \cite{choe2020attention}     & VGG16     & 30.8      \\
CONTA {[}NeurIPS20{]} \cite{zhang2020causal} & ResNet38  & 32.8      \\ 
\hline
\hline\noalign{\smallskip}
SEAM {[}CVPR20{]}  \cite{wang2020self}     & ResNet38  & 31.9      \\
DSCNet+SEAM              & ResNet38  & 34.6 \textcolor{green!80}{↑ 2.7}     \\ \hline
\hline\noalign{\smallskip}
AuxSegNet {[}ICCV21{]} \cite{xu2021leveraging} & ResNet38  & 33.9      \\
DSCNet+AuxSegNet         & ResNet38  & \textbf{35.8} \textcolor{green!80}{↑ 1.9}     \\ \hline
\hline\noalign{\smallskip}
\end{tabular}
\end{table}

\textbf{MS COCO.} To verify the generalization ability of the proposed DSCNet, we evaluated it on the more challenging MS COCO dataset. Table \ref{tab:coco} provides the segmentation performance on the COCO 2014 \textit{val} set. We observe that DSCNet outperforms SEAM and AuxSegNet by 2.7\% and 1.9\%, respectively. Figure 6 shows the visualization result, which can verify our DSCNet has a good effect on various challenging scenes.

\begin{figure}[!ht]
\centering
\includegraphics[width=3.2in]{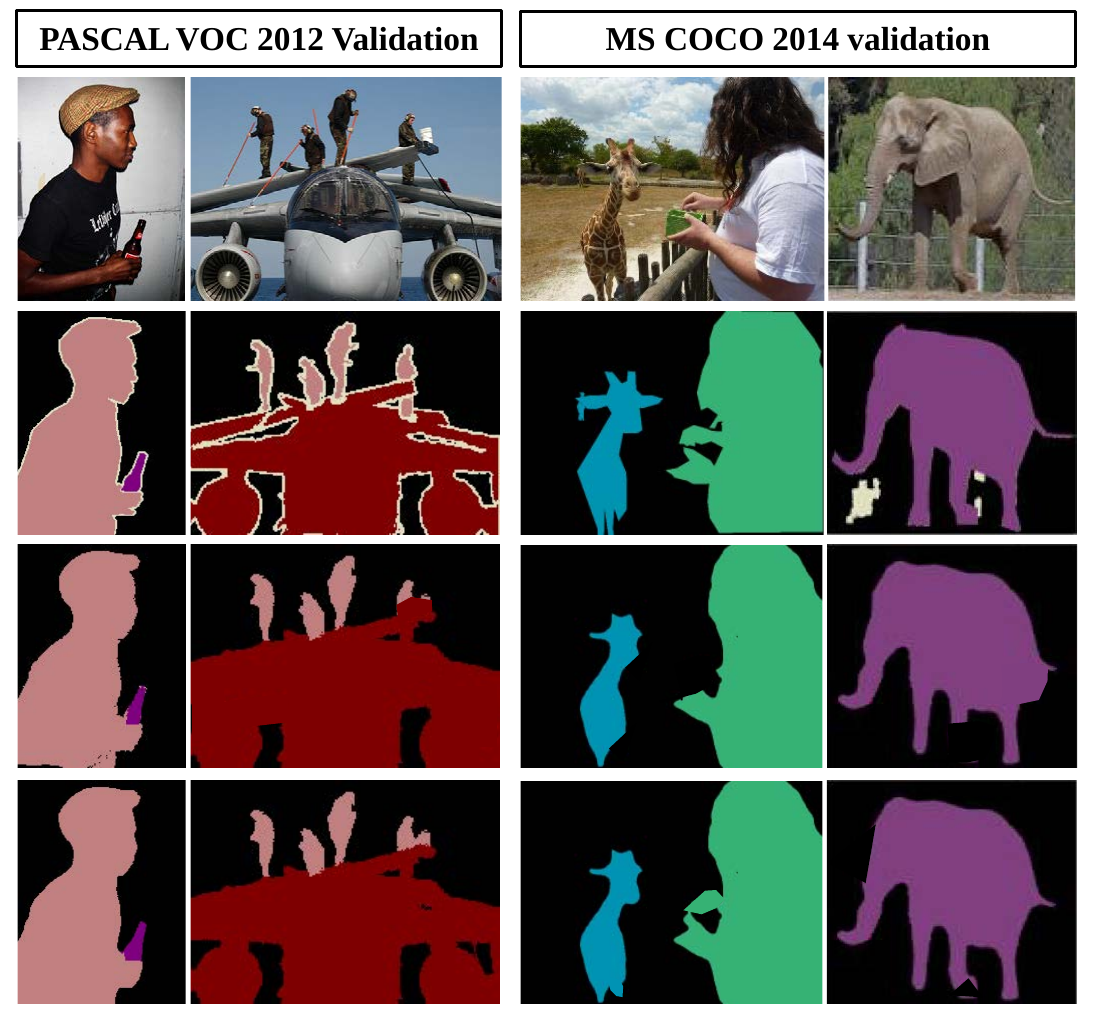}
\caption{Visualized segmentation results on VOC 2012 \textit{validation} set (left) and COCO 2014 \textit{validation} set (right). From top to bottom: input images,
ground truths, segmentation results of SEAM, as well as our DSCNet.}
\label{fig:val_result}
\end{figure}
 
%------------------------------------------------------------------------
\subsection{Ablation Study}
\label{sbusec:ablation}

In this sub-section, a comprehensive study of DSCNet is conducted through the ablation of different modules. For simplicity, we use M1, M2, M3, and M4 to respectively indicate different combinations where M1 means the baseline with the PGCL module and M2 means the baseline with the SGCL module. M3 means baseline with both two modules but consider them \textit{independently}, while M4 indicates baseline with both two modules and consider them \textit{simultaneously}. Following the previous literature [ref], the experiments are implemented on the VOC 2012 \textit{val} set. The characteristics of DSCNet through ablation on three modules are investigated in Table \ref{tab:ablative study} where mIoU $\%$ is reported. There are three significant observations:

\noindent \textbf{Each individual module is beneficial.} Comparing combinations M3 and M4 against M1$\&$M2 (\textit{only} consider pixel- or semantic-wise contextual information, we observe that each individual module is beneficial to the performance. However, the effect of SGCL is far less than that of PGCL because the SGCL tends to provide more auxiliary information to help the model construct correspondence from the semantic-level (image-level) to pixel-level.
\begin{table}[!ht]
\centering
\caption{Effectiveness of joint learning different combinations of multiple modules in terms of mIoU (\%) on PASCAL VOC 2012. For brevity, we use M1, M2, M3, and M4 to indicate different combinations, respectively. DSCL indicates a dual-stream contrastive learning mechanism. SEAM has been selected as the baseline. \label{tab:ablative study}}
\begin{tabular}{lcccc}
\hline
\hline\noalign{\smallskip}
\multirow{2}{*}{Config} & \multicolumn{2}{c}{Modules} & \multirow{2}{*}{DSCL} & \multirow{2}{*}{mIoU $(\%)$} \\ \cline{2-3}
                        & PGCL         & SGCL         &                               &                            \\ \hline
                        \hline\noalign{\smallskip}
SEAM     &              &              &                               & 64.5                       \\
+M1                  &$\surd$     &              &                               & 66.1 \textcolor{green!80}{↑ 1.6}                       \\
+M2                  &              & $\surd$      &                               & 65.4 \textcolor{green!80}{↑ 0.9}                      \\
+M3         & $\surd$      & $\surd$      &                               & 66.6 \textcolor{green!80}{↑ 2.1}                       \\
+M4                    & $\surd$     & $\surd$      & $\surd$                       & \textbf{67.7} \textcolor{green!80}{↑ 3.2}                       \\ \hline
\hline\noalign{\smallskip}
AuxSegNet     &              &              &                               & 69      \\
+M1                  & $\surd$      &              &                               & 69.6 \textcolor{green!80}{↑ 0.6}                       \\
+M2                  &              & $\surd$      &                               & 69.4 \textcolor{green!80}{↑ 0.4}                      \\
+M3         & $\surd$      & $\surd$     &      & 69.9 \textcolor{green!80}{↑ 0.9}    \\
+M4                    &$\surd$    & $\surd$      & $\surd$                       & 70.3 \textcolor{green!80}{↑ 1.3}                       \\ \hline
\hline\noalign{\smallskip}
\end{tabular}
\end{table}

\noindent \textbf{The DSCL mechanism is important.} Comparing M3 against M4, whether based on SEAM or AuxSegNet, it is found that employing DSCL mechanism achieves a large improvement (up to 1.7$\%$ in mIoU).

\noindent{\textbf{Why DSCNet does not improve much on AuxSegNet as in SEAM?}   AuxSegNet \cite{xu2021leveraging} is already an affinity learning framework and can obtain more effective information, e.g., segmentation and saliency feature maps, than SEAM \cite{wang2020self}. This verifies the importance of joint learning (i.e., Dual-Stream Contrastive Learning) for WSSS.
%------------------------------------------------------------------------
\label{sbusec:eff_ana}

\subsection{Efficiency Analysis}

Additional experiments are added in Table \ref{tab:eff_ana} to verify our pixel-wise group contrastive learning (PGCL) improves both computational cost and semantic accuracy, where "w/o PGCL" = original pixel-by-pixel modeling (i.e., no clustering). The left part of Table \ref{tab:eff_ana} denotes PGCL does not consider the time cost of clustering processing, while the right part is the time cost including clustering. Although the proposed PGCL has more components than our original pixel-by-pixel modeling module \cite{wang2020self} our method still spends less test time than it. Combined with Tables \ref{tab:voc}, \ref{tab:coco}, and \ref{tab:ablative study}, our PGCL can assist the WSSS model in obtaining much better segmentation results than the original pixel-by-pixel modeling module.
\begin{table}[!htb]
\centering
\caption{Inference performance (mIoU (\%)) and computational cost comparisons between our PGCL and original pixel-by-pixel modeling module over Pascal VOC \textit{val} split.  \label{tab:eff_ana}}
\resizebox{\linewidth}{!}{
\begin{tabular}{lcccc}
\hline
\hline\noalign{\smallskip}
\multirow{2}{*}{Baseline}& \multicolumn{2}{c}{w/o PGCL (no clustering)}  & \multicolumn{2}{c}{w/ PGCL (clustering)}   \\ \cline{2-5} 
                         & mIoU(\%) ↑  & Time (second/iter) ↓ & mIoU(\%) ↑ & Time (second/iter) ↓  \\ \hline
                         \hline\noalign{\smallskip}
SEAM                  &64.5             &1.84                  &66.1 \textcolor{green!80}{↑ 1.6}        &0.88 \textcolor{green!80}{↓ 0.96}                    \\
AuxSegNet             &69.0             &2.33                  &69.7 \textcolor{green!80}{↑ 0.7}        &1.51 \textcolor{green!80}{↓ 0.82}                   \\ \hline
\hline\noalign{\smallskip}
\end{tabular}
}
\end{table}
%\begin{table}[h]
%\centering
%\caption{Efficiency comparisons between our proposed PGCL and the original pixel-by-pixel modeling module in SEAM. \label{tab:eff_ana}}
%\begin{tabular}{lclc}
%\hline
%                            & \multicolumn{2}{c}{PGCL} & Basic \\ \hline
%Training (second/iteration) & 0.88        & 1.32       & 1.84  \\
%Testing (second/image)      & 0.75        & 1.11       & 1.66  \\ \hline
%\end{tabular}
%\end{table}

%------------------------------------------------------------------------
\subsection{Parameters Analysis}
\label{sbusec:para_ana}

\textbf{Effects of different hyper-parameter values $\alpha $ and $\beta$.} The hyper-parameter value $\alpha $ and $\beta$ in Eq. (\ref{eq:total_loss}) determines the weight of PGCL and SGCL in the segmentation task. Therefore, if we do not use PGCL or SGCL module, the DSCL mechanism can not work. For this reason, there is no result for $\alpha = 0$ or $\beta = 0$. As shown in Figure \ref{fig:parameter}, we can see that the best value of $\alpha $ and $\beta$ are 0.6 and 0.4, respectively. When $\alpha $ or $\beta$ exceeds the optimal values, mIoU basically does not improve. Compared Figure \ref{fig:parameter} (a) and (b), we can find that the effect of DSCNet on SEAM is much larger. This is because AuxSegNet itself is already the affinity learning framework and has obtained more effective information than SEAM. This verifies the importance of \textit{joint} learning (i.e., DSCL) to a certain extent.

\begin{figure}[!ht]
\begin{center}
\subfigure[SEAM]{
\includegraphics[width=1.5in,height=1in]{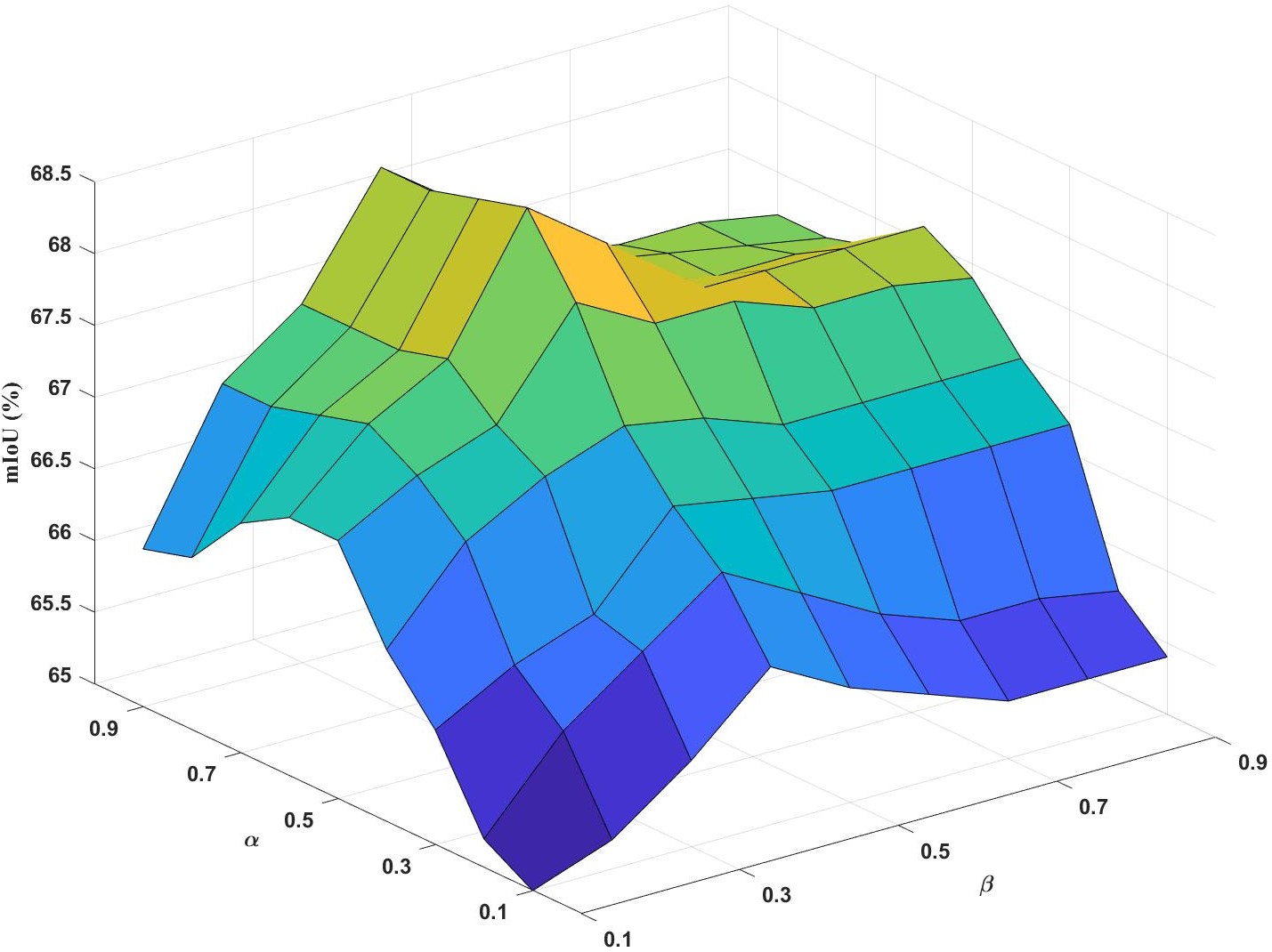}
}
\subfigure[AuxSegNet]{
\includegraphics[width=1.5in,height=1in]{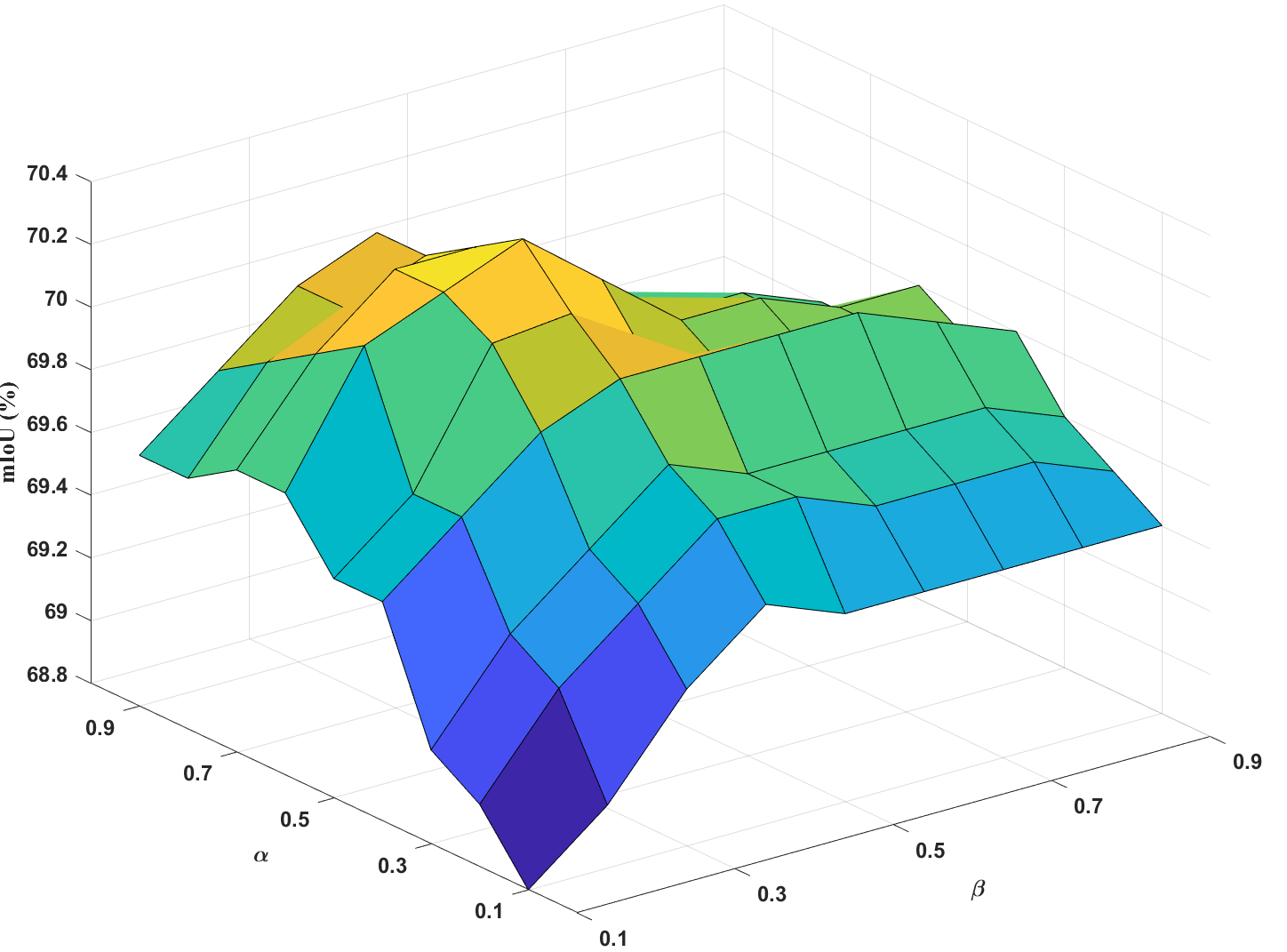}
}
\caption{Hyperparameter analysis on the PASCAL VOC 2012 \textit{val}. (a) and (b) are using SEAM and AuxSegNet as baseline, repetitively, which show the mIoU of the changing number of $\alpha$ and $\beta$.}
\label{fig:parameter}
\end{center}
\end{figure}

%-----------------------------------------------------------------------------

\section{Conclusion}
\label{sec:conclusion}
\section{Conclusion and Future Work}
\label{sec:con}

In this paper, a dual-stream end-to-end WSSS framework called DSCNet is presented which involves two novel modules: i) pixel-wise group contrastive module and ii) semantic-wise graph contrastive module. These two modules are used to extract two types of contextual information contrastive. Through a dual-stream contrastive learning mechanism, these two types of context information can be complementary to each other.
Hence, DSCNet can be endowed with the inference capacity on both pixel-wise and semantic-wise, which is beyond many existing works. Extensive experiments verify the superiority of DSCNet which improves 1.9\%$\sim$15.4\% on mIoU compared with other state-of-the-art methods. However, a drawback of our approach is that DSCNet still heavily relies on CAM maps to obtain pseudo-segmentation labels. Nevertheless, the precision of pseudo labels obtained by CAM maps is limited, which cannot sufficiently guide the subsequent WSSS task. Further exploration on pseudo labels generation under weakly supervised scenarios may further improve the performance of WSSS and are left as future work.

\begin{verbatim}


\end{verbatim}

%\bibliographystyle{IEEEtran}
%\bibliography{IEEEabrv,example}

\vspace{11pt}

\vfill

\end{document}